\documentclass[conference]{IEEEtran}

\IEEEoverridecommandlockouts
\usepackage{cite}
\usepackage{tabularx}
\usepackage{afterpage}
\usepackage{amsmath,amssymb,amsfonts}
\usepackage{algorithmic}
\usepackage{graphicx}
\usepackage{textcomp}
\usepackage{xcolor}
\def\BibTeX{{\rm B\kern-.05em{\sc i\kern-.025em b}\kern-.08em
    T\kern-.1667em\lower.7ex\hbox{E}\kern-.125emX}}
\begin{document}

\title{Machine Translation to Control Formality Features in the Target Language\\
}

\author{\IEEEauthorblockN{Harshita Tyagi}
\IEEEauthorblockA{\textit{School of Computing} \\
\textit{Dublin City University}\\
Dublin, Ireland \\
harshita.tyagi2@mail.dcu.ie}
\and
\IEEEauthorblockN{Prashasta Jung}
\IEEEauthorblockA{\textit{School of Computing} \\
\textit{Dublin City University}\\
Dublin, Ireland \\
prashasta.jung2@mail.dcu.ie}
\and
\IEEEauthorblockN{Hyowon Lee}
\IEEEauthorblockA{\textit{School of Computing} \\
\textit{Dublin City University}\\
Dublin, Ireland \\
hyowon.lee@dcu.ie}
}

\maketitle

\begin{abstract}
Formality plays a significant role in language communication, especially in low-resource languages such as Hindi, Japanese and Korean. These languages utilise formal and informal expressions to convey messages based on social contexts and relationships. When a language translation technique is used to translate from a source language that does not pertain the formality (e.g. English) to a target language that does, there is a missing information on formality that could be a challenge in producing an accurate outcome. This research explores how this issue should be resolved when machine learning methods are used to translate from English to languages with formality, using Hindi as the example data. This was done by training a bilingual model in a formality-controlled setting and comparing its performance with a pre-trained multilingual model in a similar setting. Since there are not a lot of training data with ground truth, automated annotation techniques were employed to increase the data size. The primary modeling approach involved leveraging transformer models, which have demonstrated effectiveness in various natural language processing tasks. We evaluate the official formality accuracy(ACC) by comparing the predicted masked tokens with the ground truth. This metric provides a quantitative measure of how well the translations align with the desired outputs. Our study showcases a versatile translation strategy that considers the nuances of formality in the target language, catering to diverse language communication needs and scenarios.

\end{abstract}

\section{Introduction}
In recent years, there have been significant advancements in neural machine translation (NMT), with translation quality approaching that of manual translation in various languages. However, despite these achievements, the field of controllable machine translation has received relatively little attention. While the primary focus of machine translation (MT) research has been on preserving meaning across languages, it has long been recognized by linguists and general users that maintaining pragmatic aspects of communication is also crucial[1]. 

\begin{figure}[!htbp]
\centerline{\includegraphics[width=0.35\textwidth]{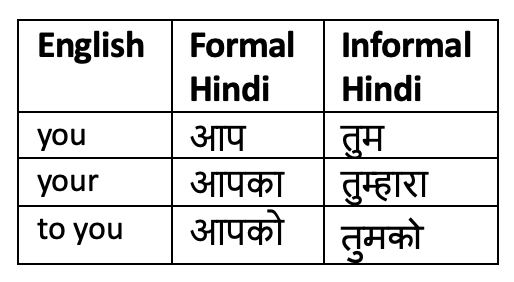}}
\caption{Examples of Formal-Informal distinction in Hindi.}
\label{fig1}
\end{figure}

Translation is a complex process that goes beyond mere linguistic conversion. One crucial aspect of translation is accounting for the variation in \emph{formality}, where language is adapted to suit specific target audiences by utilizing appropriate grammatical registers. This phenomenon is particularly evident in languages such as Japanese, Korean, Hindi, Vietnamese and Spanish, where formal and informal part-of-speech is employed to convey messages with the appropriate level of politeness and respect based on social contexts and relationships. Figure 1 demonstrates the differences between Formal Hindi and Informal Hindi translations. In the context of Indic languages, similar formality features are also present, further adding to the intricacies of translation. Each Indic language may exhibit unique variations in formality, requiring a nuanced understanding to accurately convey the intended meaning and tone in the translated text. Figure 2 depicts how a sentence translation from English to Formal Hindi and Informal Hindi works. Effectively addressing formality variations in translation is essential for ensuring cultural sensitivity, naturalness, and proper communication in the target language.

Presently accessible translators such as Google Translate primarily function in a formal translation setting, lacking the capacity to comprehend and adhere to specific grammatical registers, especially concerning formality. This limitation results in potential inaccuracies when determining the appropriate level of formality for a translation, leading to instances where the translated output may be considered inappropriate in certain contexts. In addressing this issue, several studies have attempted to control formality in MT (e.g. [2][3][4]). The concept of formality-sensitive machine translation (FSMT) was formalized, where the translation not only depends on the source segment but also on the desired target formality [5]. However, the lack of gold translations with alternate levels of formality for supervised training and evaluation has led researchers to rely on manual evaluation and synthetic supervision [6]. 

Furthermore, these studies have generally focused on adapting to formal and informal registers rather than specifically controlling grammatical formality. 

\begin{figure}[!htbp]
\centerline{\includegraphics[width=0.45\textwidth]{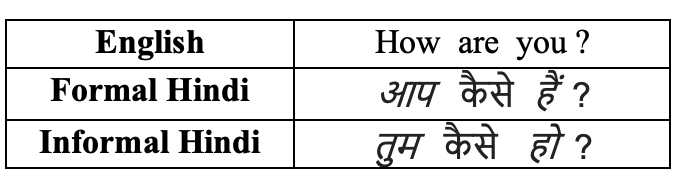}}
\caption{Example sentence of Formal-Informal distinction in Hindi.}
\label{fig2}
\end{figure}

Recognizing the importance of formality control, this research aims to develop a machine translation system that can effectively handle formality, facilitating smooth and reliable conversations and enhancing communication across languages and cultures. This system will enable more nuanced and effective linguistic exchanges, considering both formal and informal variations based on user preferences. In our research we aim to evaluate the following hypotheses. 

\vspace{10pt} %
\textbf{Hypothesis}: \textit{Pre-trained multilingual models such as IndicBERT, when fine-tuned in a formality controlled setting, will show improved performance in handling formality in machine translation compared to custom-built models for Low Resource Languages.}
\vspace{10pt} %

To create a baseline formality-agnostic translation system, we utilized the AI4Bharath’s Samanantar corpus, which is the most extensive publicly available parallel corpus for Indic languages, containing sentences paired with their English translations. Since our research focuses on understanding and extracting formality features from the target language, we identified additional data from the IWSLT2022 repository. This additional corpus allowed us to perform automated annotation and labelling to augment the formality features within a larger training set. By incorporating this additional data, we aimed to enhance the system’s ability to capture the nuances of formality during the translation process.

In this paper, we compare the performance of custom-built and fine-tuned bilingual models with formality-controlled tuning on pre-trained multilingual model such as IndicBERT, which is a Masked Language Model (MLM). We evaluate the official formality accuracy (ACC) by comparing the predicted masked tokens with the ground truth. The outputs are first detokenized and detruecased, using the 13A Moses detokenizer. Additionally, we calculate and compare the Cross-Entropy-Loss throughout the pre-training and fine-tunning process. For evaluation, we have a set of hypotheses (H), along with corresponding formal references (F) and informal references (I), and we use a function to obtain phrase level contrastive terms from a reference. Finally, we aim to showcase a real-time translation interface capable of translating English to Hindi, taking into consideration the formality feature based on the user’s preference.

\section{Related Work}

Several advancements in NMT have been made, in addition to the transformer. These include domain adaptation, back-translation, Bidirectional and Auto-Regressive Transformers (BART), multilingual de-noising pre-training for NMT (mBART), and Generative Adversarial Networks (GAN). While these techniques have shown considerable improvements for High-Resource Languages (HRLs), which are language pairs with an abundant amount of training data (such as English (EN) $\leftrightarrow$ German (DE) or English (EN) $\leftrightarrow$ French (FR)), they fall short in achieving state-of-the-art performance for Low-Resource Languages (LRLs). Insufficient improvements in the LRLs corpus can be attributed to the lack of data, which impedes the learning of proper sentence context. The literature to this research intends to explore the state-of-the-art translation models and leverage formality control features to increase model performance on Low Resource Languages.

\subsection{Custom-built Bi-lingual Translation Model}

One of the key objectives is to design and develop a bilingual model specifically tailored to perform effectively in Low-Resource Languages. The inspiration was drawn from a research by Sonali et al. [7], proposed to implement CNN based encoder and RNN decoder. The current NMT works perfectly fine until it faces sentences with longer structure, with issues such as lack of robustness due to rare words, slow inference speed, and the inability to translate all words that persisted in vanilla RNN-NMT and made it unsuitable. Google attempted to address these shortcomings by employing an 8-layered LSTM RNN system and was successful in potentially reducing 60 percent of translation errors. An encoder, as CNN can simultaneously encode source sentence, captures relationships between element sequences. CNN has some advantages over RNN due to their intricate structure, which allows them to perform parallel computation, increasing training speed and easily removing the vanishing gradient problem. These RNN decoder variants have gates in their cells that decide what information to keep and what to discard during model training, allowing them to handle longer sequences of sentences more effectively.

Another study figured out that in technical terms, the basic encoder-decoder model encounters an information bottleneck issue [8]. This occurs when the model must compress all the information of the source language into a fixed-length vector, without knowing the length of the input sentence. Consequently, the translation model struggles to handle longer sentences. However, an attention architecture provides a solution to this problem. At each step of the decoder, a connection is established to the encoder, allowing the model to focus on specific parts of the source sentence. This connection produces a weighted sum of encoder values that are used in the decoder part to enhance translation quality. 

To minimize the vanishing gradient problem and to handle long-term dependencies, GRU (Gated Recurrent Units) sets the current memory value based on the previous memory value, throughout all time-steps in the deep neural network. This approach makes attention-based GRU model the optimal choice over the basic RNN algorithm, also confirmed by a comparative study by Ajinkya et al. [9]. An attention-based GRU Seq2Seq predictor was combined with a Transitional Matrix-based predictor. An artificial test dataset with predefined transitional states is used to present and validate hard attention (Vivekanandh et al. [10]). This experiment lead to 10-12 percent gain in prediction accuracy, compared to the conventional attention models.

\subsection{Automated Annotation}

After establishing the baseline model, our next step was to tackle the challenge of limited formality annotations in the training corpus. To address this issue, we examined previous research studies for guidance. Kusum Lata et al. [11] proposed a rule-based mention detection method followed by post editing, adopting the Begin Inside-Outside (BIO) format to annotate mentions in Hindi text. This approach focused on accurately identifying formality markers in the text. In another study by Danial Zhang et al. [12], a language-specific post-editing strategy was employed to achieve optimal performance. To overcome the scarcity of formality annotations, they suggested developing a formality classifier that uses weakly labelled data augmentation. This technique generates synthetic formality labels from a large parallel corpus, enabling the augmentation of training data. 

Rippeth et al. [13] introduced the concept of synthetic supervision, which involves using synthetic training samples as a replacement or complement to expensive paired contrastive samples. They proposed automatically annotating naturally occurring bitext to identify formality, thus generating formal and informal samples. This approach prioritizes control over the formality markers in the output, albeit with less supervision than paired contrastive samples. By leveraging rule-based mention detection, language-specific post-editing, weakly labelled data augmentation, and synthetic supervision, we aimed to enhance the training process and improve the accuracy of formality control in our model.

\subsection{Formality Control}

Since the ultimate goal of this study is to achieve control over formality features in language translation, the primary objective is to investigate the existing state-of-the-art methods and formality tuning techniques that have shown promising results in high-resource languages and determine their applicability in low-resource languages. A study by Vincent et al. [14] used a pre-trained model, fine-tuned it on the supervised data to control the desired formality of the hypothesis with a tagging approach by Sennrich et al. [2] whereby a formality-indicating tag is appended to the source input. This method has been widely used in research in various controlling tasks (e.g. [15][16][17]). 

A recent study by Priyesh Vakharia et al. [18] looked at previous works incorporating contrasting styles [13][19] as motivation for controlling styles, using an additive intervention approach. This approach entails adding a single style intervention vector V to the pre-trained encoder output Z. The same vector V is added to all the tokens of the encoder outputs, thereby changing the encoder outputs uniformly. They modified the above approach to allow for more flexibility while learning. Instead of a single intervention vector V, they propose a unique vector Vi for every token i in the input space, in short to repurpose an Embedding layer as a style intervening layer between the encoder and the decoder.

\section{Methodology}

\subsection{Data Collection}\label{AA}

\subsubsection{IWSLT2022 Corpus}\label{AA}
To facilitate the task of formality-controlled translation, we utilized the dataset provided by the International Workshop on Spoken Language Translation (IWSLT) 2022. The dataset consists of a parallel corpus comprising 400 sentence pairs for English-to-Hindi translations, with explicit formality annotations. The main structure of the IWSLT2022 data corpus is depicted in Figure 3. Given the relatively small size of the corpus and the low-resource nature of Hindi, we will leverage the formality features extracted from this dataset to augment formality in a larger training corpus automatically.

\begin{figure}[!htbp]
\centerline{\includegraphics[width=0.5\textwidth]{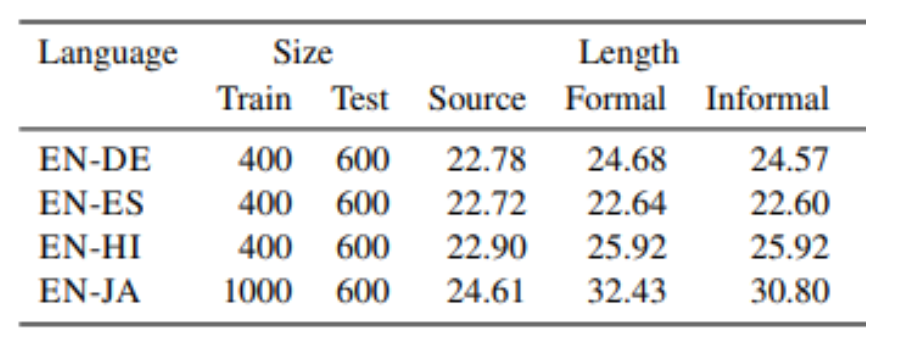}}
\caption{IWSLT2022 data corpus}
\label{fig3}
\end{figure}

\subsubsection{Samanantar Corpus}\label{AA}
Additionally, we obtained a substantial dataset of 10 million sentences from the Indian Institute of Technology Madras (IIT-M) called the Samanantar Corpus [20]. This corpus encompasses a diverse range of sentence sources, including phone conversations, news headlines, and TED talks, providing a rich variety of linguistic patterns and contexts. Initially, we employed this dataset to train a formality-agnostic bilingual model capable of accurately translating between English and Hindi. Subsequently, we utilized the formality features derived from the IWSLT2022 corpus to annotate the Samanantar Corpus. This annotation process enabled us to train a formality control model, allowing users to adjust the formality of the translated text while preserving translation quality.

We use a random split of 0.2 to construct the validation dataset during model development.

\begin{figure}[!htbp]
\centerline{\includegraphics[width=0.45\textwidth]{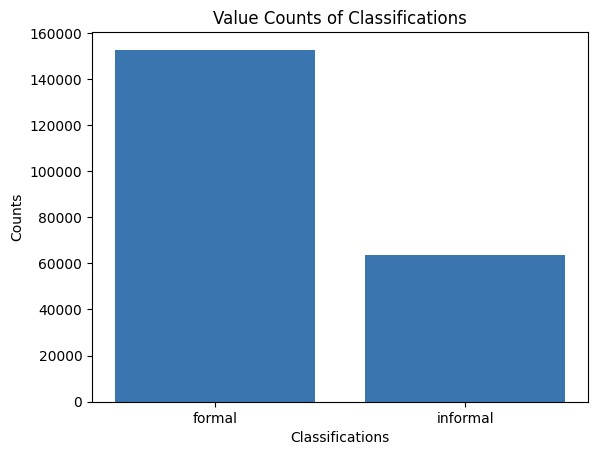}}
\caption{Formal vs. Informal words}
\label{fig4}
\end{figure}

\subsection{Data Pre-Processing}\label{AA}
In the data pre-processing phase, various techniques were employed to prepare the input data for the subsequent translation model training. The first step involved tokenization of the Hindi and English texts, where sentences were split into tokens, and only the initial 100 tokens were considered for further analysis. To ensure data standardisation and cleaning, a custom standardisation function was applied. This function transformed the input strings to lowercase and removed punctuation marks and special characters using regular expressions. 

The vocabulary size was then determined by counting the unique tokens in both languages. Text vectorization was carried out using the TensorFlow Keras TextVectorization layer, which facilitated the conversion of the text data into numeric representations.

Additionally, a specific standardisation function was applied during vectorization for the Hindi text, addressing additional cleaning requirements. The dataset was appropriately formatted, including the truncation of decoder inputs by removing the last token. This process resulted in a well-prepared dataset that was then used for model training, ensuring the optimal representation of the input data and facilitating subsequent translation tasks.

\subsection{Automated Data Labelling}\label{AA}
Given our reliance on labelled data, we adopted a strategic approach to expand our training corpus by following the methodology proposed in the work  at IWSLT 2022 conference [14]. To achieve this, we leveraged the IWSLT2022 corpus to extract words that had been previously annotated. Upon extraction, we separated the words into two distinct sets: one for formal language and another for informal language. This segregation allowed us to create a clear distinction between the two linguistic styles.

To automate the labelling process, we developed a script designed to identify the extracted words within the Samantar corpus. The script then accurately labeled each word as either ``Formal'' or ``Informal'' based on its respective set. This automated approach significantly expedited the annotation process and ensured consistency in labeling. This assumption was grounded in the notion that the presence of a significant number of formal or informal words within a sentence influenced its overall linguistic style. To make the annotations explicit and easily identifiable within the Samantar corpus, we adopted a specific annotation format. Whenever a word from the formal set was found, we enclosed it within annotation tags [F] and [/F]. Likewise, when an informal word was detected, we encapsulated it within [I] and [/I] tags. This tagging system allowed us to maintain the integrity of the original sentences while highlighting the formal and informal elements.

The underlying assumption we made during this process was that sentences containing words from the formal set would be considered formal sentences, while sentences containing words from the informal set would be classified as informal sentences. We labelled the sentences which do not contain both formal and informal words as ``Neutral''. Figure 4 depicts the classification-based distribution of sentences in the dataset. The two basic classes are shown, 'formal' and 'informal,' with the 'neutral' category eliminated from this analysis. The graph depicts the relative frequency of formal and informal sentences, suggesting that formal sentences outnumber informal sentences in the sample.

By undertaking this process, we successfully expanded and enriched our training corpus with labelled data, which in turn enhanced the performance and accuracy of our language model for formal and informal language understanding.

\subsection{Modelling}\label{AA}

\subsubsection {Bilingual Model}\label{AA}

In this part of the research, a bilingual model for translation is developed using a combination of various techniques and architectures. The aim is to address the limitations observed in existing NMT models and achieve improved performance in both high resource and low-resource language settings. To build the bilingual model, a transformer-based architecture is employed as the foundation [22]. The transformer model has demonstrated significant advancements in NMT and serves as a starting point. However, additional enhancements are introduced to tackle specific challenges.

The model consists of two primary components: the Transformer Encoder and the Transformer Decoder. The Transformer Encoder incorporates a multi-head attention mechanism, which allows the model to capture relationships between different elements in the source sentence effectively. It further utilizes dense projections and layer normalization techniques to enhance the learning process. The Encoder is responsible for encoding the source language sentence and extracting meaningful representations from it. On the other hand, the Transformer Decoder focuses on generating the target language translation. It utilizes three multi-head attention mechanisms: the first attends to the input sentence itself, enabling the model to capture the context within the target sentence. The second attention mechanism attends to the output of the Encoder, facilitating the alignment of the source and target sentences, and the third, which enables the model to learn underlying patterns occurring due to the formality tagged sentences. The Decoder also incorporates dense projections and layer normalization for better modeling of the translation process.

To incorporate positional information, a Positional Embedding layer is employed for both the Encoder and Decoder. This layer combines token embeddings, representing the words in the input sentence, with positional embeddings that encode the position of each word in the sequence. This enables the model to understand the sequential nature of the sentence and capture positional relationships. The entire model is trained using a large corpus of bilingual data. During training, attention masks are used to handle padding and enforce causal attention, ensuring that the model attends only to the previous positions during decoding.

By combining the power of the transformer architecture, attention mechanisms, dense projections, and positional embeddings, the proposed bilingual model attempts to achieve state-of-the-art translation performance. The research addresses the challenges faced in both high-resource and low-resource language scenarios, leveraging advancements in NMT to build a robust and effective translation system, while also controlling formality features.

\subsubsection {Multilingual Model}\label{AA}

In this study, we opted to leverage pre-trained model IndicBERT v2, which is a Masked Langauge Model (MLM) instead of building a multilingual model from scratch. These models were fine-tuned using formality control settings to cater to our specific requirements. IndicBERT v2, a multilingual BERT model [22], was trained on the extensive IndicCorpv2 dataset, providing comprehensive coverage for 24 Indic languages. Notably, IndicBERT v2 exhibits remarkable performance, comparable to strong baselines, and outperforms other models across 7 out of 9 tasks on the IndicXTREME benchmark. This underscores the efficacy and versatility of IndicBERT v2 in addressing a wide array of natural language processing tasks in diverse Indic languages.

The Masked language models (MLMs) are the type of language model that is trained on a dataset of text that has been masked out. This means that some of the words in the text have been replaced with special tokens, such as [MASK]. The MLM is then trained to predict the original words that were masked out. MLMs are a powerful tool for natural language processing tasks, such as text generation, machine translation, and question answering. They are able to learn the meaning of words and phrases by seeing how they are used in context.

\subsection{Training \& Fine Tuning}\label{AA}
\subsubsection{Bilingual Model}\label{AA} 

For optimization and fine-tuning process, Optuna is implemented to fine-tune the hyperparameters of the bi-lingual Transformer model, specifically for the task of sequence-to-sequence translation. The objective is to maximize the accuracy metric on a validation set by tuning various hyperparameters of the model. The key hyperparameters being optimized are sequencelength, batchsize, embeddim, latentdim, and numheads.

The optimization process begins by defining an objective function that takes in a set of hyperparameters sampled by Optuna. Within this function, a Transformer model is constructed with an encoder-decoder architecture using Keras. The encoder and decoder components of the Transformer are built, with specific hyperparameters such as the embedding dimension, number of attention heads, and latent dimension being determined based on the sampled values from the trial.

The model is then compiled with an appropriate optimizer (RMSprop) and loss function (sparse categorical cross-entropy). The model is trained for 30 epochs and then evaluated on the validation set to obtain the accuracy metric. Optuna's study function is used to create an optimization study, specifying that the objective is to maximize the accuracy. The optimization process is carried out, and the best hyperparameters and corresponding accuracy metric are recorded.

\subsubsection{Multilingual Model}\label{AA}
We focus on fine-tuning the AI4Bharath/IndicBERTv2-MLM model for formality control. Formality control is a crucial aspect of natural language processing that aims to modify the level of formality in sentences, catering to different language styles and contexts. The IndicBERTv2-MLM model is a state-of-the-art language model pre-trained on a large corpus of Hindi text using masked language modeling (MLM). Our goal is to adapt this powerful pre-trained model to better understand and generate formal and informal sentences.

To achieve this, we created a custom dataset comprising 1000 sentences in Hindi, with a balanced mix of formal and informal expressions. We tokenized and preprocessed the data, applying text cleaning techniques to ensure consistency and reduce noise. The dataset is then split into training and testing sets for evaluating model performance.

We initialized the IndicBERTv2-MLM model and attached a language modeling head for masked language modeling fine-tuning. We employed the Adam optimizer and set the learning rate to 1e-4 to optimize the model during training. Instead of employing a constant learning rate during the training process, we opted for a dynamic approach using a learning rate scheduler. The purpose of this scheduler is to adjust the learning rate dynamically throughout training, which aids the model in converging faster while avoiding issues like overshooting or being trapped in local minima. To achieve this, we utilized the linear scheduler with warm-up function from the transformers library to construct a learning rate scheduler with warm-up steps. This scheduler gradually increases the learning rate from a very low value to the desired learning rate, which is particularly beneficial in the initial stages of training, leading to more stable and effective. 

We iteratively train the model over 30 epochs, optimizing the parameters to minimize the cross-entropy loss. The model’s progress and loss values are monitored through each epoch to assess training efficacy. After every 30 epochs, a new set of 1000 records are trained again with the same epochs to introduce fresh data for the model to learn from. This process is known as ``restarting'' or ``re-batching'' the training. By using a new set of records, we ensure that the model encounters a diverse range of examples in each batch, leading to better generalization and reducing the risk of overfitting. Additionally, this approach helps to mitigate any potential bias that may have been introduced due to the order of the data during initial training.

During the fine-tuning process, we leveraged the power of random token masking to simulate formal-informal variations in the training data. By randomly masking tokens within sentences, we enhance the model's ability to generalize and make contextually appropriate predictions. Additionally, we created masked versions of the ground truth sentences and evaluate the model's loss on these masked tokens to assess its formality control capabilities. Upon completing the fine-tuning, we test the model on a separate test dataset to evaluate its performance on unseen data. The model's ability to generate formality-controlled outputs for new sentences demonstrates its utility in various language applications.

\subsection{Evaluation}\label{AA}

Upon loading the fine-tuned IndicBERTv2-MLM model, which has been trained to perform masked language modeling, we applied a basic text cleaning function to preprocess the dataset, converting the text to lowercase, removing special characters, and stripping extra white spaces. The data is then split into a sample of 100 records for evaluation purposes.

Next, we created a custom evaluation function, which takes the model, tokenizer, data loader, and device as input. During evaluation, we set the model to evaluation mode to ensure that the model does not update its parameters during inference. We then generated translations for the masked sentences using the generate method with the maxlength, numbeams, and earlystopping parameters to control the translation process. The generated translations are decoded into Hindi sentences using the tokenizer, and the results are collected into a list of translated sentences.

Before evaluating the masked language model's performance, we created a copy of the original formality annotated column and applied random masking to the sentences while preserving the tags. The masking probability is set to 0.15, and sentences are masked inside tags with this probability. The masked sentences are then tokenized using the tokenizer, and predictions are generated for the masked tokens using the fine-tuned model. We also used the MosesDetokenizer to detokenize the predicted tokens into complete Hindi sentences for evaluation.

In the evaluation phase, we calculated the number of correct predictions for masked tokens and the total number of tokens in the ground truth sentences. By comparing the predicted tokens with the ground truth tokens, we calculated the accuracy of the model's predictions. Given a set of hypotheses H, sets of corresponding phrase-annotated formal references F and informal I, and a function $\Phi$ yielding phrase-level contrastive terms from a reference, the task-specific evaluation metric is defined as follows:

\[
\text{match}_{\underset{f}{}} = \sum_{j} [\Phi(\text{F}_{\underset{j}{}})\in\text{H}_{\underset{j}{}}^ {\Phi(\text{I}_{\underset{j}{}})\notin\text{H}_{\underset{j}{}})}]
\]

\[
\text{match}_{\underset{i}{}} = \sum_{j} [\Phi(\text{F}_{\underset{j}{}})\notin\text{H}_{\underset{j}{}}^ {\Phi(\text{I}_{\underset{j}{}})\in\text{H}_{\underset{j}{}})}]
\]

\[
\text{acc}_{\underset{j}{}} = \text{match}_{\underset{j}{}} / (\text{match}_{\underset{f}{}} + \text{match}_{\underset{i}{}} ), j \in {f,i}
\]

We note that the task accuracy is a function of the number of matches in the hypotheses, not the number of expected phrases, i.e. $match_f + match_{if} \leq |H|$ [13].

Finally, we calculated the loss obtained during evaluation and the accuracy of the model's masked predictions. The loss represents the model's performance in terms of cross-entropy on the masked tokens in the ground truth sentences, while the accuracy (ACC) indicates the percentage of correctly predicted tokens compared to the total number of tokens in the ground truth sentences. This evaluation plan enables us to assess the effectiveness of the fine-tuned masked language model for formality control on the given dataset.

\begin{figure*}[t]
    \centering
    \includegraphics[width=\textwidth]{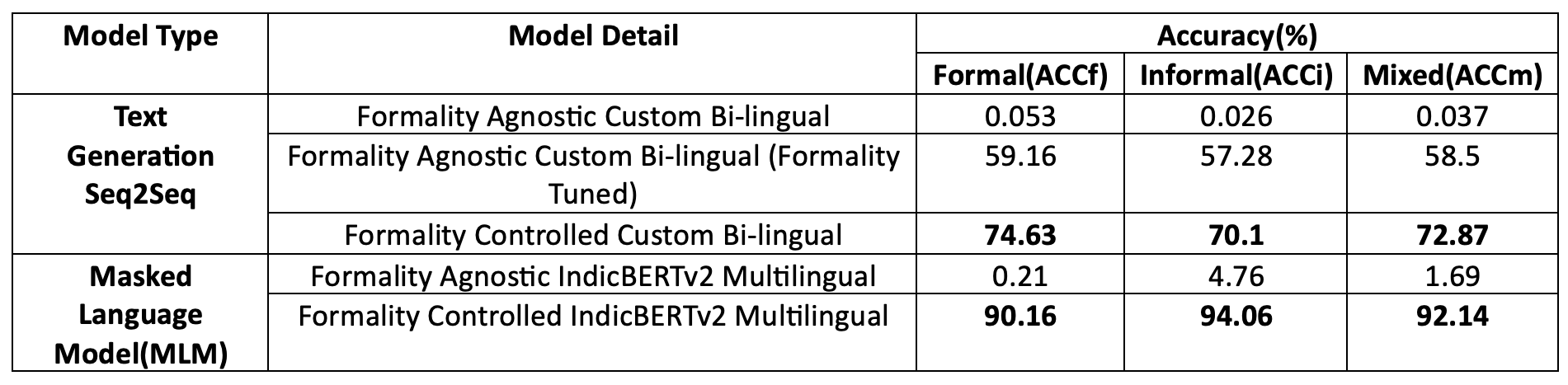}
    \caption{Evaluation Results}
    \label{fig5}
\end{figure*}

\section{Challenges}

\subsection{Lack of Training Data}\label{AA}
Formality-sensitive languages usually lacks sufficient parallel corpora that is labelled for training successful machine translation models, especially for the Low Resource Languages (LRL's). To address this obstacle, we implemented an automation technique to label and annotate the minimal training data available. We were able to augment the dataset and construct a more extensive training corpus.  As a result of this strategy we improved the model's capacity to handle different levels of formality. 

\subsection{Lack of Support Material or Existing Code Base}\label{AA}
We discovered the scarcity of freely available support material or existing code bases focusing explicitly on formality control in machine translation during our investigation. To overcome this difficulty, we did a thorough literature analysis and communicated with the machine translation expert to learn about current methodologies who suggested to focus more on formality characteristics in bilingual models, as formality norms in multi-lingual models would be different for different languages, resulting in lower accuracy and to investigate pre-trained models and compare them to the custom bilingual model. Based on this understanding, we created bespoke code to implement and test various formality control mechanisms. Drawing from this understanding, we leveraged customized code generators like BlackBoxAI, ChatGPT and BART to formulate and experiment with various formality control mechanisms. Furthermore, we intend to publicly share our code and discoveries with the community in order to add to the body of knowledge in this subject.

\subsection{Lack of Computing Resources}\label{AA}
One of the primary obstacles we encountered throughout this project was the limited computing resources at our disposal for constructing a formality-tuned bilingual model trained on extensive datasets, which could rival the existing large pre-trained models. Moreover, fine-tuning large text generation pre-trained models like mBART \& mT5 was challenging, leading us to switch to masked language models like IndicBERT. Despite this challenge, we managed to leverage the GPU resources available on the Google Colab platform. Although it did not entirely eliminate the resource constraints, it significantly enhanced our ability to train and explore more complex models compared to relying solely on CPU-based machines.

\section{Results}

We present the performance of our IndicBERTv2 model over multiple epochs during the fine tuning process in Figure 6. Loss measures the discrepancy between the predicted values of our model and the actual ground truth labels. As the training progresses, we aim to minimize this loss, indicating the model is effectively learning to make accurate predictions. At the beginning of the training process (epoch 10), the loss value stands at 0.72. As the model becomes more familiar with the data and refines its internal parameters, the loss steadily decreases. By epoch 20, the loss has already diminished to 0.544, reflecting the model's improved performance. 

\begin{figure}[!htbp]
\centerline{\includegraphics[width=0.45\textwidth]{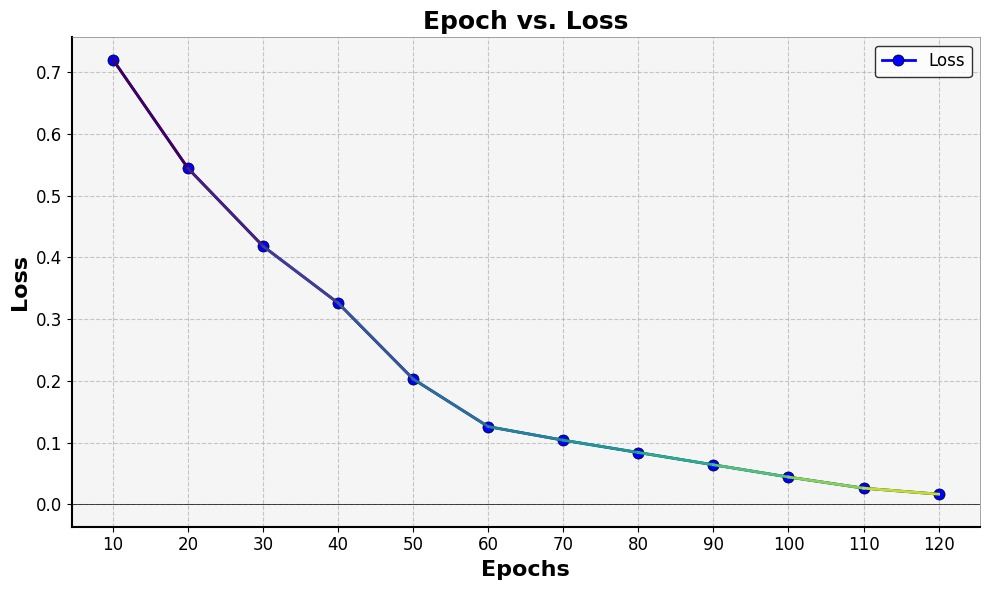}}
\caption{Loss vs. Epochs}
\label{fig6}
\end{figure}

As training continues, the model continues to fine-tune its parameters and adapt to the underlying patterns in the data. By epoch 70, it drops to 0.104, indicating that the model is now highly proficient at capturing complex patterns in the data and making precise predictions As we progress to higher epochs, the loss continues to decrease. Towards the end of the training process, the model's performance continues to improve. At epoch 120, the model attains an outstandingly low loss of 0.0167. This performance is a result of the model's capacity to adapt to various data patterns and generalize across different instances.

For the same model, the plot in Figure 7 showcases the performance trend of the model across different epochs. The line plot exhibits a smooth curve with markers representing the accuracy at specific epochs. As the number of epochs increases, the model's accuracy significantly improves, indicating the effectiveness of the training process.

\begin{figure}[!htbp]
\centerline{\includegraphics[width=0.45\textwidth]{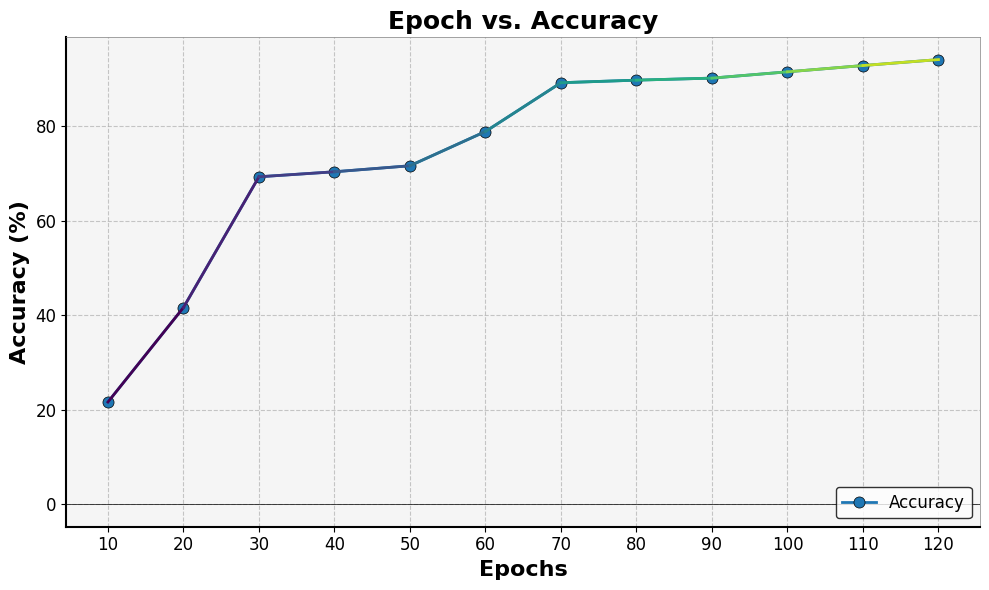}}
\caption{Accuracy vs. Epochs}
\label{fig7}
\end{figure}

We categorically present the outcomes of our training and fine-tuning experiments on various models for formality agnostic and formality controlled bilingual natural language processing tasks in Figure 5. Our primary objective was to develop sophisticated models capable of handling formality variations in both formal and informal languages. The evaluation metrics used for our analysis are Accuracy (\%) scores, reflecting the performance of the models on different datasets.

For text generation tasks using Seq2Seq models, we evaluated two variations of our formality agnostic custom bi-lingual model. In the first experiment, which we refer to as ``Formality Agnostic Custom Bi-lingual'', the model achieved low accuracy, with 5.3\% accuracy for formal language and 2.6\% accuracy for informal language. In the second experiment, where we fine-tuned the same custom bi-lingual model with formality tuning, we observed remarkable improvements, achieving an impressive 59.16\% accuracy for formal language and 57.28\% accuracy for informal language. This substantial enhancement in performance demonstrates the effectiveness of incorporating formality tuning into our model.

Additionally, we introduced a formality controlled custom bi-lingual model, specifically designed to handle formality variations from the start. This model achieved superior results compared to the formality agnostic versions, attaining an accuracy of 74.63\% for formal language and 70.1\% for informal language. These outcomes highlight the significance of formality control in developing advanced language models that cater to diverse linguistic contexts.

Furthermore, we explored Masked Language Models (MLM) using the IndicBERTv2 Multilingual model, both in formality agnostic and formality controlled setups. In the formality agnostic experiment, the MLM model demonstrated low accuracy, with only 0.31\% for formal language and 4.76\% for informal language. However, upon applying formality control to the same model, we achieved substantial improvements, with accuracy soaring to 90.16\% for formal language and an impressive 94.06\% for informal language.

These results highlight the significance of formality-aware models in multilingual natural language processing tasks. By incorporating formality tuning and formality control, we have observed substantial gains in accuracy, enabling the models to successfully adapt to both formal and informal language inputs. These findings showcases the potential of our approach to be a valuable asset for applications in language generation, translation, and other language-related tasks in various linguistic settings.

\subsection{Masked Language Models (MLM) Vs Text Generation Models}
These are two prominent approaches in natural language processing tasks that serve distinct purposes. MLMs, like BERT and GPT-3, are pre-trained models designed to understand language patterns and context by predicting masked words within a sentence. They are trained on vast amounts of unlabeled data from diverse sources, making them adept at capturing intricate language relationships. The key advantage of MLMs lies in their ability to handle language variations, as they learn from the patterns in the data directly. This flexibility makes MLMs well-suited for multilingual applications, where formality variations and diverse linguistic contexts are prevalent.

On the other hand, Text Generation Models, such as Seq2Seq models, are designed explicitly for generating human-like text. They learn to map input sequences to output sequences and can be fine-tuned for specific tasks with labeled data. Text Generation Models shine when fine-tuned with formality annotations, allowing them to produce contextually appropriate language with a desired level of formality. They excel in applications where precise control over generated text is required, such as language translation, chatbots, and text summarization.

While MLMs are versatile models that can adapt to various language variations without requiring explicit formality annotations, leveraging vast unlabeled data, Text Generation Models offer fine-grained control over the formality of the generated text by utilizing labeled data for specific tasks. The choice between these approaches depends on the availability of annotated data, the level of formality control needed, and the linguistic diversity of the target applications. In our study, we leveraged both models along with formality control annotations to enhance their performance in identifying nuances in diverse language contexts.

\section{Conclusions}

Our study demonstrates the successful adoption and efficacy of formality control strategies in machine translation for Low-Resource Languages. Returning to our hypothesis stated in the Introduction section, we can now confirm that multilingual models outperform bilingual models in formality control settings, particularly for English to Hindi translation. The observed improvements during the training process highlight the potential for enhancing bilingual models with sufficient resources.

Moving forward, we recognize the importance of gathering a larger and more diversified dataset that includes both formal and informal translations, which can significantly enhance machine translation in formality-sensitive languages. Such data will facilitate the development of contextually accurate and culturally sensitive translations. Fine-grained formality control now becomes feasible, offering better alignment with desired levels of formality in various situations and scenarios.

Our study opens up avenues for deeper scientific analysis of formality transfer strategies and language variation challenges. By applying these findings to multilingual applications, we can  foster linguistic inclusivity and enable successful cross-lingual communication. We believe our work serves as a foundation for future research in formality-aware machine translation systems, encouraging the prioritization and sharing of tagged datasets to advance the field further.


\begin{thebibliography}{00}
\bibitem{b1} Eduard Hendrik Hovy. 1987. Generating natural language under pragmatic constraints. Ph.D. Dissertation. Yale University, USA. Order Number: AAI8729079.

\bibitem{b2} Rico Sennrich, Barry Haddow, and Alexandra Birch. 2016. Controlling Politeness in Neural Machine Translation via Side Constraints. In Proceedings of the 2016 Conference of the North American Chapter of the Association for Computational Linguistics: Human Language Technologies, pages 35–40, San Diego, California. Association for Computational Linguistics.

\bibitem{b3} Weston Feely, Eva Hasler, and Adrià de Gispert. 2019. Controlling Japanese Honorifics in English-to-Japanese Neural Machine Translation. In Proceedings of the 6th Workshop on Asian Translation, pages 45–53, Hong Kong, China. Association for Computational Linguistics.

\bibitem{b4} Andrea Schioppa, David Vilar, Artem Sokolov, and Katja Filippova. 2021. Controlling Machine Translation for Multiple Attributes with Additive Interventions. In Proceedings of the 2021 Conference on Empirical Methods in Natural Language Processing, pages 6676–6696, Online and Punta Cana, Dominican Republic. Association for Computational Linguistics.

\bibitem{b5} Xing Niu, Sudha Rao, and Marine Carpuat. 2018. Multi-Task Neural Models for Translating Between Styles Within and Across Languages. In Proceedings of the 27th International Conference on Computational Linguistics, pages 1008–1021, Santa Fe, New Mexico, USA. Association for Computational Linguistics.

\bibitem{b6} Niu, Xing, and Marine Carpuat 2020. Controlling neural machine translation formality with synthetic supervision. Proceedings of the AAAI Conference on Artificial Intelligence. Vol. 34. No. 05. 2020.

\bibitem{b7} S. Sharma, M. Diwakar, P. Singh, A. Tripathi, C. Arya and S. Singh 2021. A Review of Neural Machine Translation based on Deep learning techniques. 2021 IEEE 8th Uttar Pradesh Section International Conference on Electrical, Electronics and Computer Engineering (UPCON), Dehradun, India, 2021, pp. 1-5, doi: 10.1109/UPCON52273.2021.9667560.

\bibitem{b8} J. Singh, S. Sharma and B. J 2022. Natural Language Processing based Machine Translation for Hindi-English using GRU and Attention. 2022 International Conference on Applied Artificial Intelligence and Computing (ICAAIC), Salem, India, 2022, pp. 965-969, doi: 10.1109/ICAAIC53929.2022.9793214

\bibitem{b9} A. Naik, M. Karani, K. Chheda and M. Gada 2022. NMT for English-Marathi using RNNs \& Attention. 2022 6th International Conference On Computing, Communication, Control And Automation (ICCUBEA, Pune, India, 2022, pp. 1-6, doi: 10.1109/ICCUBEA54992.2022.10010782

\bibitem{b10} V. Elangovan, W. Xiang and S. Liu 2023. A RealTime C-V2X Beamforming Selector Based on Effective Sequence to Sequence Prediction Model Using Transitional Matrix Hard Attention, in IEEE Access, vol. 11, pp. 10954- 10965, 2023, doi: 10.1109/ACCESS.2023.3241130.

\bibitem{b11} Lata, K., Singh, P. \& Dutta, K 2023. Semi-automatic Annotation for Mentions in Hindi Text. SN COMPUT. SCI. 4, 515 (). https://doi.org/10.1007/s42979-023-01885-z

\bibitem{b12} Daniel Zhang, Jiang Yu, Pragati Verma, Ashwinkumar Ganesan, and Sarah Campbell. 2022. Improving Machine Translation Formality Control with Weakly-Labelled Data Augmentation and Post Editing Strategies. In Proceedings of the 19th International Conference on Spoken Language Translation (IWSLT 2022), pages 351–360, Dublin, Ireland (in-person and online). Association for Computational Linguistics.

\bibitem{b13} Elijah Rippeth, Sweta Agrawal, and Marine Carpuat. 2022. Controlling Translation Formality Using Pre-trained Multilingual Language Models. In Proceedings of the 19th International Conference on Spoken Language Translation (IWSLT 2022), pages 327–340, Dublin, Ireland (in-person and online). Association for Computational Linguistics.

\bibitem{b14} Sebastian Vincent, Loïc Barrault, and Carolina Scarton. 2022. Controlling Formality in Low-Resource NMT with Domain Adaptation and Re-Ranking: SLT-CDT-UoS at IWSLT2022. In Proceedings of the 19th International Conference on Spoken Language Translation (IWSLT 2022), pages 341–350, Dublin, Ireland (in-person and online). Association for Computational Linguistics.

\bibitem{b15} Melvin Johnson, Mike Schuster, Quoc V. Le, Maxim Krikun, Yonghui Wu, Zhifeng Chen, Nikhil Thorat, Fernanda Viégas, Martin Wattenberg, Greg Corrado, Macduff Hughes, and Jeffrey Dean. 2017. Google’s Multilingual Neural Machine Translation System: Enabling Zero-Shot Translation. Transactions of the Association for Computational Linguistics, 5:339– 351.

\bibitem{b16} Eva Vanmassenhove, Christian Hardmeier, and Andy Way. 2018. Getting gender right in neural machine translation. Proceedings of the 2018 Conference on Empirical Methods in Natural Language Processing, EMNLP 2018, pages 3003–3008.
 
\bibitem{b17} Surafel Melaku Lakew, Mattia Di Gangi, and Marcello Federico. 2019. Controlling the Output Length of Neural Machine Translation. In Proceedings of the 16th International Conference on Spoken Language Translation, Hong Kong. Association for Computational Linguistics.

\bibitem{b18} Priyesh Vakharia, Shree Vignesh S, and Pranjali Basmatkar. 2023. Low-Resource Formality Controlled NMT Using Pre-trained LM. In Proceedings of the 20th International Conference on Spoken Language Translation (IWSLT 2023), pages 321–329, Toronto, Canada (in-person and online). Association for Computational Linguistics.

\bibitem{b19} Andrea Schioppa, David Vilar, Artem Sokolov, and Katja Filippova. 2021b. Controlling machine translation for multiple attributes with additive interventions. In Proceedings of the 2021 Conference on Empirical Methods in Natural Language Processing, pages 6676–6696, Online and Punta Cana, Dominican Republic. Association for Computational Linguistics

\bibitem{b20} Gowtham Ramesh, Sumanth Doddapaneni, Aravinth Bheemaraj, Mayank Jobanputra, Raghavan AK, Ajitesh Sharma, Sujit Sahoo, Harshita Diddee, Mahalakshmi J, Divyanshu Kakwani, Navneet Kumar, Aswin Pradeep, Srihari Nagaraj, Kumar Deepak, Vivek Raghavan, Anoop Kunchukuttan, Pratyush Kumar, and Mitesh Shantadevi Khapra. 2022. Samanantar: The Largest Publicly Available Parallel Corpora Collection for 11 Indic Languages. Transactions of the Association for Computational Linguistics, 10:145–162.

\bibitem{b21} Sumanth Doddapaneni, Rahul Aralikatte, Gowtham Ramesh, Shreya Goyal, Mitesh M. Khapra, Anoop Kunchukuttan, and Pratyush Kumar. 2023. Towards Leaving No Indic Language Behind: Building Monolingual Corpora, Benchmark and Models for Indic Languages. In Proceedings of the 61st Annual Meeting of the Association for Computational Linguistics (Volume 1: Long Papers), pages 12402–12426, Toronto, Canada. Association for Computational Linguistics.

\bibitem{b22} A. Vaswani, N. Shazeer, N. Parmar, J. Uszkoreit, L. Jones, A. N. Gomez, L. Kaiser, and I. Polosukhin 2017, Attention is All You Need, in Advances in Neural Information Processing Systems 30 (NIPS 2017), Long Beach, CA, USA, 2017.

\end{thebibliography}
\end{document}